\documentclass[runningheads]{llncs}
\usepackage{graphicx}
\usepackage{amsmath,amssymb} 
\usepackage{bm}
\usepackage{color}
\usepackage{caption}
\usepackage[width=122mm,left=12mm,paperwidth=146mm,height=193mm,top=12mm,paperheight=217mm]{geometry}

\usepackage{comment}
\usepackage{xargs}                      
\usepackage[pdftex,dvipsnames]{xcolor}  
\newcommandx{\unsure}[2][1=]{\todo[linecolor=red,backgroundcolor=red!25,bordercolor=red,#1]{#2}}

\begin{document}
\pagestyle{headings}
\mainmatter
\def\ECCV18SubNumber{2}  

\title{DeeSIL: Deep-Shallow Incremental Learning} 



\author{Eden Belouadah, Adrian Popescu}
\institute{CEA, LIST, Vision and Content Engineering Lab, F-91191 Gif-sur-Yvette, France\\
\{eden.belouadah,adrian.popescu\}@cea.fr}

\maketitle

\begin{abstract}
Incremental Learning (IL) is an interesting AI problem when the algorithm is assumed to work on a budget. 
This is especially true when IL is modeled using a deep learning approach, where two complex challenges arise due to limited memory, which induces catastrophic forgetting and delays related to the retraining needed in order to incorporate new classes. 
Here we introduce $DeeSIL$, an adaptation of a known transfer learning scheme that combines a fixed deep representation used as feature extractor and learning independent shallow classifiers to increase recognition capacity. 
This scheme tackles the two aforementioned challenges since it works well with a limited memory budget and each new concept can be added within a minute. 
Moreover, since no deep retraining is needed when the model is incremented, $DeeSIL$ can integrate larger amounts of initial data that provide more transferable features. 
Performance is evaluated on ImageNet LSVRC 2012 against three state of the art algorithms. 
Results show that, at scale, $DeeSIL$ performance is 23 and 33 points higher than the best baseline when using the same and more initial data respectively. 

\end{abstract}

\section{Introduction and background}
Typical deep learning pipelines are well adapted to solve tasks when all training data is available at all times and there are loose constraints regarding time available for training. 
Under these conditions, augmenting the classification ability can simply be done by learning a new representation, either from scratch or via Fine Tuning (FT). 
However, when one or both of the above conditions are violated, adding new classes becomes non-trivial. 
The authors of iCaRL~\cite{DBLP:conf/cvpr/RebuffiKSL17} rightfully note that there exists no satisfactory algorithm that can qualify as class-incremental. They frame three necessary properties of it: (i) be trainable from new stream data that occurs arbitrarily; (ii) provide competitive performance for past classes when new ones are integrated and (iii) computational requirements and memory footprint should remain bounded.  

In iCaRL, recognition capacity is incremented by retraining for every new batch of classes. 
A fixed-size memory is used to store positive examples which provide a compact approximate representation of known classes. 
For each new batch of classes, iCaRL starts with updating its representation by adding all available data for the new classes to known examples.
After each state the fixed memory is updated with examples from the newly learned classes.
To counter catastrophic forgetting~\cite{mccloskey:catastrophic}, \textit{i.e.} the tendency of a neural net to forget old information when new information is ingested, classification and distillation losses are used.
While fulfilling the three necessary conditions for a class-incremental algorithm, the performance reduction is still important since top-5 accuracy drops from roughly 90\% for 100 to 45\% for 1000 classes~\cite{DBLP:conf/cvpr/RebuffiKSL17}.

Learning-without-Forgetting (LwF)~\cite{DBLP:conf/eccv/LiH16} combines knowledge distillation and Fine Tuning. The authors first perform a warm-up step by optimizing new parameters only, then the whole network is optimized using classification loss for new tasks and distillation loss for old tasks. 
A LwF adaptation for IL is introduced in~\cite{DBLP:conf/cvpr/RebuffiKSL17} and has the advantage of not requiring a memory for past data. 
However, its performance is lower than that of iCaRL in a single task scenario. 
Aljundi et al.~\cite{DBLP:conf/cvpr/AljundiCT17} introduced ExpertGate, an architecture based on a network of experts from which only the most adapted one is activated. 
A gating mechanism is applied to training samples to decide which expert to transfer knowledge from. 
When a new task arrives, a new expert is added and knowledge is transferred from previous models using FT or LwF~\cite{DBLP:conf/eccv/LiH16}. 
Expert Gate learns a good data representation when augmenting the number of tasks. However, it violates the third property of IL algorithms since its number of parameters increases with the number of tasks.
The authors of~\cite{DBLP:conf/cvpr/WangRH17} and~\cite{DBLP:journals/corr/RusuRDSKKPH16} improve the plasticity of deep architectures by widening existing layers and/or deepening the network. 
While this improves recognition ability, the drawback in a constrained setting is that the number of parameters is increased when augmenting the network's capacity.

We introduce $DeeSIL$, an adaptation of a known transfer learning scheme~\cite{DBLP:conf/cvpr/RazavianASC14,ginsca15semfeat,DBLP:journals/corr/abs-1805-08974} to incremental learning. 
In order to qualify as class-incremental and maximize flexibility, $DeeSIL$ includes two weakly correlated steps. 
First, a deep model is trained in order to provide fixed representations which are then used to learn independent shallow classifiers during the incremental phase. 
Instead of using the system memory to keep positive examples, a set of negative features that are necessary to train classifiers incrementally is stored. 
This choice makes it possible to use all positive examples for training without violating the memory constraint. 
Our hypothesis is that independent shallow learning over all positives compensates the drawback related to the use of a fixed deep representation.
Since no deep retraining is needed to increase system capacity, the approach is considerably less complex compared to its purely deep learning counterparts. 
The addition of a new class is done through the training of a shallow classifier, an operation that takes less than a minute on a single CPU.
$DeeSIL$ is tested against three competitive IL algorithms, including iCaRL~\cite{DBLP:conf/cvpr/RebuffiKSL17}, the best such algorithm known to the authors. 
The ImageNet LSVRC 2012 dataset is used for evaluation and results show significant improvement for the proposed method. 

\section{Method} 
An overview of $DeeSIL$ is provided in Fig. 1.  
The algorithm is an adaptation for incremental learning of a well-known transfer learning scheme~\cite{DBLP:journals/corr/abs-1805-08974}. 
Given a set of images $X^i$ for a class to be learned, features $F^i$ are extracted using a fixed deep representation provided by the deep features extractor ($\pmb{DFE}$). 
Then a shallow binary classifier $\pmb{C^i}$ is trained using $F^i$ as positives and $F^N$ as negatives in order to predict the activations $p^i$ of the class for test images. 
$F^N$ is the memory of the system and it contains a constant number $K$ of features, regardless of the state of the system (\textit{i.e.} number of recognizable classes). 
$F^N$ is generated by the negative selector ($\pmb{NS}$) component which is the main adaptation introduced in $DeeSIL$ to make a classical transfer learning pipeline~\cite{DBLP:journals/corr/abs-1805-08974} suitable for incremental learning. 
Given \textbf{A} ($y$ recognizable classes), the initial state of the system, the following steps are needed to move to state \textbf{B} ($y+j$ classes): 
(1) extract features for the $j$ new classes; (2) update the pool of negatives $F^N$ using $\pmb{NS}$ component; (3) train $j$ shallow classifiers. Following common practice in transfer learning~\cite{DBLP:journals/corr/abs-1805-08974,DBLP:conf/cvpr/RazavianASC14,ginsca15semfeat}, we use linear SVMs. We further discuss steps (1) and (2) hereafter.  
 \begin{figure} 
 \label{fig:deesil-overview}
 \centering
\includegraphics[width=0.83\columnwidth]{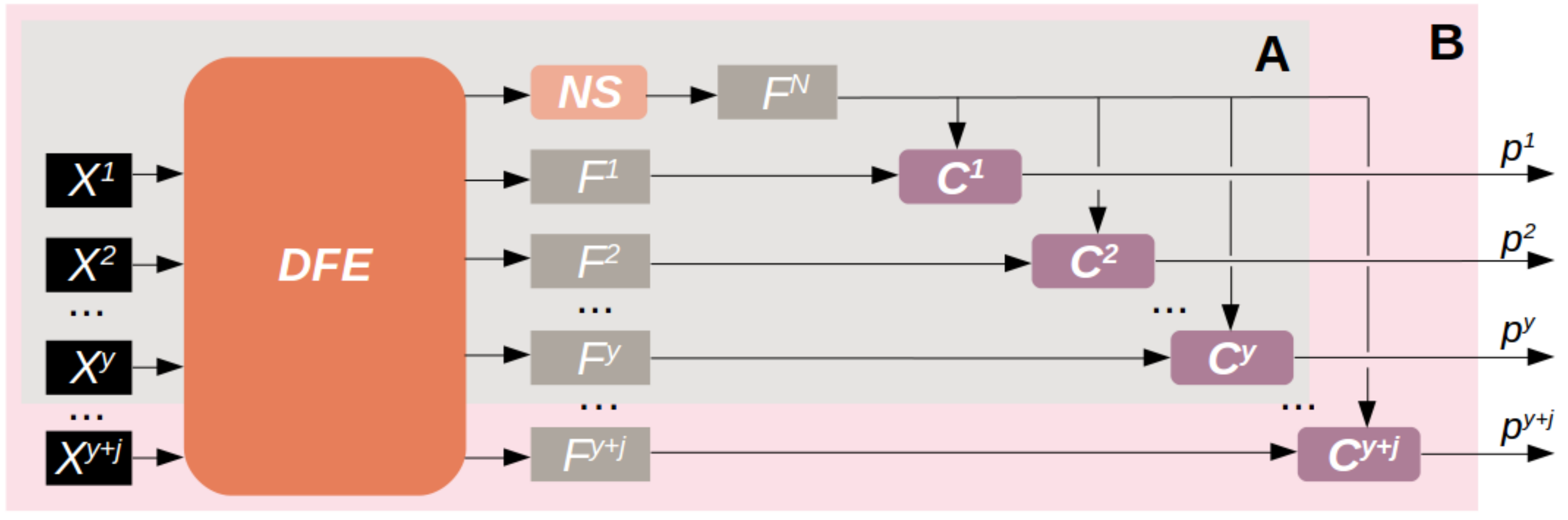}
 \caption{Overview of $DeeSIL$. Two states of the system, \textbf{A} (light gray background) and \textbf{B} (light pink background) that recognize respectively $y$ and $y+j$ classes are presented. $X^{i}$, $F^{i}$ are sets of images and features for the $i^{th}$ class. $F^N$ is a set of negative features obtained using a negative selector ($\pmb{NS}$) and common to all shallow classifiers that are added in a given state. $\pmb{DFE}$ is a deep features extractor. $\pmb{C^i}$ is a shallow classifier learned for the $i^{th}$ class and the output $p^i$ is the associated prediction.}
 \end{figure}

\textbf{Deep features extractor.}
In~\cite{DBLP:conf/cvpr/RebuffiKSL17}, each new state of the class-incremental-algorithm depends on the representation learned in the preceding state.
Here, deep features extraction and shallow classifier learning are separated. 
$DeeSIL$ thus implements a form of transfer learning which uses a fixed deep representation. 
To evaluate the effect of the amount of training data and its visual proximity with the test data, we train three variants of $\pmb{DFE}$:
\begin{itemize}
\item $IN100$ - train only with the ImageNet data of the initial state, a setting that is directly comparable with~\cite{DBLP:conf/cvpr/RebuffiKSL17};
\item $IN1000$ - train with a larger dataset that has similar characteristics with the test set but has no common classes with it. 
One thousand classes are selected to form a diversified subset of ImageNet and thus increase universality (\textit{i.e.} optimize their transferability toward new tasks)~\cite{tamaazousti2017mucale,tamaazousti2018universal}. 
\item $FL1000$ - train with a more challenging dataset which is obtained from weakly annotated Flickr group data and is visually more distant from the test set. 
Within each group, a semi-supervised reranking~\cite{Chen:2015:WSL:2919332.2920124} is initially performed to remove a part of noisy images.
\end{itemize}
A greedy algorithm~\cite{deselaers2009} which operates with classes' mean representations is used for dataset diversification in the last two variants.  
It picks at each iteration the class which is on average least similar to those already selected. 
Visual representations from $IN100$ are used as basis for the diversification process.
\\\\ 
\textbf{Negatives selection.}
In standard transfer learning~\cite{DBLP:journals/corr/abs-1805-08974}, shallow classifiers are learned in a one-VS-rest fashion since all data is available at all times. 
Here, a selection is necessary to fit $F^N$ features in the memory budget $K$ for any state of the algorithm. 
We test three negative selection strategies:
\begin{itemize}
\item $ind$ - following~\cite{ginsca15semfeat} $F^N$ is composed of $K$ YFCC image features~\cite{Thomee:2016:YND:2886013.2812802} selected so as to represent frequent but diversified tags.  
\item $rand$ - a random and balanced sampling of image features from all past and current classes. 
\item $div$ - diversified samples from all recognizable classes. The greedy algorithm implemented for dataset diversification is reused here at image level. 
\end{itemize}
For $rand$ and $div$, if DeeSIL recognizes $y$ classes in a given state, each class will have $\frac{K}{y}$ representatives in $F^N$.
Naturally, a class' own representatives are discarded from $F^N$ when training its shallow classifier.

\section{Evaluation and discussion}
$DeeSIL$ is tested using the ILSVRC 2012  dataset~\cite{Russakovsky:2015:ILS:2846547.2846559}.
The evaluation protocol (order of classes, size of system states) is nearly identical to the one used for iCaRL~\cite{DBLP:conf/cvpr/RebuffiKSL17}. 
ILSVRC 2012 includes a total of 1000 classes, further split into 10 batches of 100 classes, which means that 10 distinct states of the class-incremental algorithms are tested.
The test set is the same but, since we need to optimize the SVMs, we keep out 20 images for validation and train on remaining images. 
We use the best three systems from~\cite{DBLP:conf/cvpr/RebuffiKSL17} as baselines: (1) iCaRL - their contribution and the best IL algorithm known to us; (2) LwF-MC - adaptation of Learning without Forgetting ~\cite{DBLP:conf/eccv/LiH16} to IL scenario and (3) Fixed Representation - training over a frozen initial network, except for the classification layer.  

ResNet-18~\cite{DBLP:conf/cvpr/HeZRS16} was trained from scratch using PyTorch~\cite{paszke2017automatic} following the methodology described in~\cite{DBLP:conf/cvpr/HeZRS16} with 100 and 1000 ImageNet classes and 1000 Flickr groups. 
Training images are processed using a random resized crop of size $224\times224$ and a random horizontal flip and they are normalized after these transformations.  
An SGD optimizer is used. 
The learning rate starts at 0.1 and is divided by 10 when the error plateaus for 10 consecutive epochs. The weight decay is 0.0001 and the momentum is 0.9.
Each configuration is trained for 100 epochs and the model with optimal accuracy is retained. 
The penultimate layer (average pooling with 512 dimensions) is extracted by $\pmb{DFE}$ and then L2-normalized before being fed into the shallow classifiers. The memory size, which stores negative features, is $K=20000$, the same as in~\cite{DBLP:conf/cvpr/RebuffiKSL17}.

The SVM classifiers were optimized with 20 images per class for validation. 
Values of the regularization parameter between $0.0001$ and $1000$ were tried and the optimal parameter was then used in each variant of the system.  
 
\begin{figure}
\centering
\begin{minipage}{.5\textwidth}
  \centering 
  \captionsetup{width=.9\linewidth}
  \includegraphics[width=.999\linewidth]{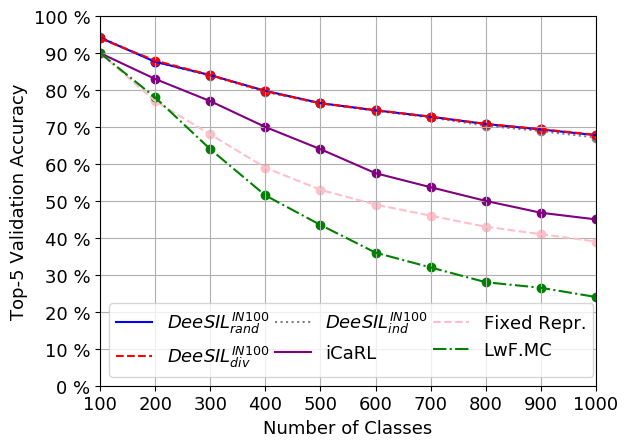}
  \captionof{figure}{Top-5 accuracy on ILSVRC for $DeeSIL$ variants obtained with three negative selection strategies.}
  \label{fig:gauche}
\end{minipage}%
\begin{minipage}{.5\textwidth}
  \centering
    \captionsetup{width=.9\linewidth}
  \includegraphics[width=.999\linewidth]{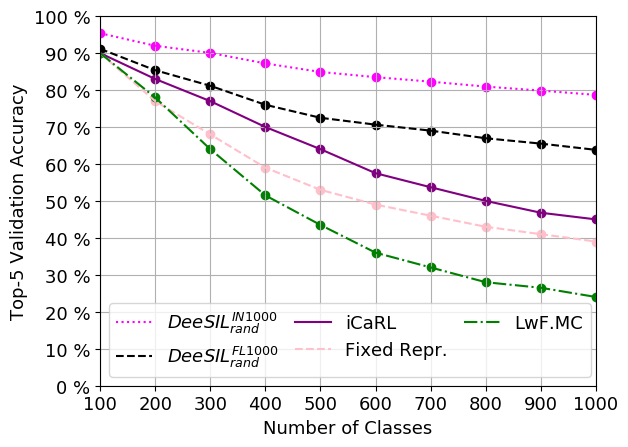}
  \captionof{figure}{Top-5 accuracy on ILSVRC for fixed deep representations obtained with larger datasets.}
  \label{fig:droite}
\end{minipage}
\end{figure}

The results in Fig.~\ref{fig:gauche} show that all variants of $DeeSIL$, trained with $rand$, $div$ and $ind$ negatives selection outperform the state of the art systems. 
At scale, \textit{i.e.} 1000 classes learned incrementally, performance increases from 45\% for iCaRL to 68\% when $rand$ and $div$ negatives are exploited.
This gain is consistent over all the states of the class-incremental evaluation, with larger difference for large batches.
$DeeSIL$ can be seen as a variant of Fixed Representation learning but differs from it through the use of all positives in the incremental phase.
This leads to an even higher performance gain than in the case of iCaRL.

The three $\pmb{NS}$ variants have rather performance and this finding shows that our method is robust w.r.t. the choice of negatives. 
Selecting negatives from the dataset ($rand$ and $div$) gives marginally better results ($0.5$ points gain) compared to the use of an independent negative set ($ind$) for 1000 classes.
${rand}$ being simpler to compute than $div$, $DeeSIL_{rand}$ will be used in further experiments.

In Fig.~\ref{fig:droite}, we test the effect of using more data to obtain strong fixed representations. 1000 ImageNet classes and Flickr groups are used respectively.  
Richer data compensates for the fact that features are transferred from classes that are different from the tested ILSVRC classes. 
This is especially the case for $DeeSIL_{rand}^{IN1000}$, which exploits a subset of ImageNet distinct from ILSVRC.
Performance improvements of 10 and 33 points are obtained over $DeeSIL_{rand}^{IN100}$, the best configuration trained with the 100 initial ILSVRC classes and over iCaRL respectively. 
$DeeSIL_{rand}^{FL1000}$, the version trained on non-curated Flickr data has lower performance than $DeeSIL_{rand}^{IN1000}$, but is close to $DeeSIL_{rand}^{IN100}$ and still well above the state of the art algorithms. 
The last result confirms the finding in~\cite{Chen:2015:WSL:2919332.2920124} that it is possible to learn reasonable representations even with little or no manually data. 

Beyond performance, it is also important to compare the complexity of $DeeSIL$ to that of iCaRL, the main baseline. 
ResNet-18, the basic deep architecture is the same for both methods. 
Recognition capacity incrementation is done with linear SVMs. 
This entails the computation of a dot product per class, which is equivalent to adding a class in the final layer of a CNN.

Training is simpler in $DeeSIL$ since a single deep network training is needed at the beginning. 
In the incremental step, we only train shallow classifiers. 
Adding a single class typically takes less than 1 minute, distributed among deep features extraction and SVM training on an INTEL-Xeon-E5-2650-v2@2.60GHz CPU. 
For comparison, adding a batch of 100 new classifiers in iCaRL takes approximately 32 hours on an NVIDIA Titan X GPU. 
Incremental learning is typically needed in low-resource contexts and, assuming that an initial deep representation is available, $DeeSIL$ can be deployed even in absence of a GPU.
Equally important, due to the independent learning of shallow classifiers, $DeeSIL$ can seamlessly  integrate batches of new classes of arbitrary size. 
In contrast, purely deep learning based algorithms need retraining and this step is particularly long if one class is added at a time. 

Compared to iCaRL, the focus is shifted from positive to negative selection to fill in the memory of the system. 
As shown in the experiments, our algorithm is affected by catastrophic forgetting to a much lesser extent. 
The choice to select negatives is beneficial for scalability in terms of number of learnable classes. 
Given a memory budget $K$, iCaRL can learn at most $y \leqslant K$ classes while $DeeSIL$ can learn as many classes as presented to the system.
Naturally, negatives selection becomes more complicated if $y \geqslant K$ since not all known classes will be represented anymore. 
Also, while the same number of items is stored in iCaRL and $DeeSIL$, memory needs are lower in our case since we store $512$ dimensional features instead of images of past classes.

It is interesting to evaluate the decrease in performance compared to a situation in which all training data is available at all times. 
ResNet-18~\cite{DBLP:conf/cvpr/HeZRS16} top-5 accuracy on 1000 ILSVRC classes trained with all data is approximately 89\%. iCaRL halves this score while our best configurations with $\pmb{DFE}$ based on 100 and 1000 classes lose only 22 and 12 points respectively. 
The gap could probably be further reduced if the feature extractors were more universal~\cite{tamaazousti2017mucale,tamaazousti2018universal}. 
This could, for instance, be achieved if $DeeSIL$'s initial training would be done with an even larger number of classes. 

\section{Conclusion}
We revisit a known transfer learning scheme for it to fit the three necessary conditions needed to qualify as a class-incremental algorithm~\cite{DBLP:conf/cvpr/RebuffiKSL17}.
The proposed method achieves significantly better performance than existing algorithms while also being much faster to train and more scalable in terms of number of learnable classes.
To facilitate reproducibility, classifier configurations and data used to train them will be made public.
The results presented here encourage us to pursue the development of $DeeSIL$ along the following directions : (1) test the effect of using lower size memory, (2) push the evaluation to a much larger number of classes to test the limits of the different methods and (3) integrate more universal deep representations to improve overall performance.

\clearpage

\bibliographystyle{splncs}
\bibliography{egbib}
\end{document}